\documentclass[11pt]{article}
\DeclareFontFamily{U}{mathb}{}
\DeclareFontShape{U}{mathb}{m}{n}{
   <5> <6> <7> <8> <9> <10> gen * mathb
   <10.95> mathb10
   <12> mathb12
   <14.4> mathb12
   <17.28> mathb17
   <20.74> mathb17
   <24.88> mathb17
}{}
\DeclareSymbolFont{mathb}{U}{mathb}{m}{n}
\DeclareMathSymbol{\dlsh}{3}{mathb}{"EA}
\newcommand{\ttnl}{$\dlsh$\\}

\usepackage[preprint]{acl}

\usepackage[nott]{newtxtext}
\usepackage{newtxmath}

\usepackage{relsize}
\usepackage{CJKutf8}
\NewDocumentCommand{\ctext}{m}{\begin{CJK}{UTF8}{gbsn}\smaller{#1}\end{CJK}} 

\usepackage{calc}
\allowdisplaybreaks 

\newcommand{\narrowtt}[1]{\texttt{#1}\hspace{-0.4em}} 
\usepackage{booktabs}
\usepackage{tabularx}
\usepackage{multirow}

\defcitealias{bncconsortium_2007_british}{BNC Consortium} 
\newcommand{\citealpaliasyear}[1]{\citetalias{#1}, \citeyear{#1}}

\usepackage{latexsym}

\usepackage[T1]{fontenc}

\usepackage[utf8]{inputenc}

\usepackage{microtype}

\usepackage{inconsolata}

\usepackage{graphicx}

\title{
    Sakura at BEA 2026 Shared Task 1: What Makes Vocabulary Difficult?
    }

\author{%
\normalfont%
\textbf{Adam Nohejl}\textsuperscript{1}\quad
\textbf{Xuanxin Wu}\textsuperscript{2}\quad
\textbf{Yusuke Ide}\textsuperscript{3}\\
\textbf{Maria Angelica Riera Machin}\textsuperscript{3}\quad
\textbf{Yi-Ning Chang}\textsuperscript{4}\quad
\textbf{Hitomi Yanaka}\textsuperscript{1,5,6}%
\\
\\
 \textsuperscript{1}RIKEN\quad
 \textsuperscript{2}The University of Osaka\quad
 \textsuperscript{3}Nara Institute of Science and Technology\\
 \textsuperscript{4}National Tsing Hua University\quad
 \textsuperscript{5}The University of Tokyo\quad
 \textsuperscript{6}Tohoku University
\\
\vphantom{A}\parbox[t]{\textwidth}{\centering
    \small\texttt{adam.nohejl@riken.jp}\quad
    \texttt{xuanxin.wu@ist.osaka-u.ac.jp}\quad
    \texttt{ide.yusuke.ja6@is.naist.jp}\\[1pt]
    \small\texttt{riera\_machin.maria.rn9@naist.ac.jp}\quad
    \texttt{changyn@gapp.nthu.edu.tw}\quad
    \texttt{hyanaka@is.s.u-tokyo.ac.jp}}
}

\begin{document}
\maketitle
\begin{abstract}
We describe two types of models for vocabulary difficulty prediction: a high-accuracy black-box model, which achieved the top shared task result in the open track, and an explainable model, which outperforms a fine-tuned encoder baseline. As the black-box model, we fine-tuned an LLM using a soft-target loss function for effective application to the rating task, achieving $r > 0.91$. The explainable model provides insights into what impacts the difficulty of each item while maintaining a strong correlation ($r > 0.77$). We further analyze the results, demonstrating that the difficulty of items in the British Council's Knowledge-based Vocabulary Lists (KVL) is often affected by spelling difficulty or the construction of the test items, in addition to the genuine production difficulty of the words. We make our code available online.\footnote{\resizebox{\linewidth - 1.8em}{!}{\url{https://github.com/ynklab/vocabulary-difficulty}}}

\end{abstract}

\section{Introduction}

The goal of the BEA 2026 Shared Task on Vocabulary Difficulty Prediction for English Learners \citep{bea2026st-findings} is to build models of the difficulty of English words given a learner's L1. Such difficulty predictions can be used as a basis for pedagogical materials or computer-adaptive tests.

The shared task utilizes a large dataset with vocabulary difficulty scores for L1 Chinese, German, and Spanish learners, spanning thousands of vocabulary test items, the British Council's Knowledge-based Vocabulary Lists (KVL; \citealp{schmitt_etal_2021_introducing,schmitt_etal_2024_knowledge}).
Each test item, as shown in the example in \autoref{tab:kvl-ex}, consists of an equivalent L1 word, L1 context, and a clue for the first letter of the English word and its length in letters.

The test format, therefore, focuses on productive knowledge, in particular, on the ability to write the English word with the correct spelling given the test prompt. The difficulty scores, which we aim to predict, are intercept values of a generalized linear mixed model (GLMM), i.e., the log-odds of a learner responding correctly.

\begin{table}[tb]
\sffamily\smaller%
\renewcommand{\arraystretch}{1.2}%
\setlength{\tabcolsep}{4pt}
\begin{tabularx}{\linewidth}{|l|l|X|}
\hline
\multicolumn{2}{|l|}{L1} & Spanish\\
\hline
\multicolumn{2}{|l|}{English word} & house\\
\hline
\multicolumn{2}{|l|}{Part of speech} & noun\\
\hline
\multirow{4}{*}{\rotatebox[origin=c]{90}{Test item}} & L1 word & casa\\
\cline{2-3}
& L1 context & Vivo en una casa grande que tiene tres dormitorios.\\
\cline{2-3}
& Clue & h \_ \_ \_ \_\hfill\color{gray}first letter and blanks\\
\hline
\multicolumn{2}{|l|}{Difficulty score} & 3.07\hfill\color{gray}$\uparrow$ easy, $\downarrow$ difficult\\
\hline\end{tabularx}
\caption{Example of an item in the KVL data.}\label{tab:kvl-ex}
\end{table}

The shared task consisted of a closed track and an open track. While the open track was completely unrestricted, in the closed track, 
the use of large language models (LLMs) was limited to feature extraction, and only the L1-specific training data could be used. We present two approaches to vocabulary difficulty modeling in this setting, each leveraging LLMs in a different way:

\paragraph{LLMs fine-tuned using soft targets.} We propose a simple yet novel technique to fine-tune LLMs for continuous value prediction using soft targets. Models based on this technique have outperformed all other shared task submissions in the open track.

\paragraph{Explainable model.} We build a model using well-defined features, such as similarity to the L1 word and spelling difficulty, some of which are based on LLM prompting, and use SHapley Additive exPlanations (SHAP; \citealp{lundberg_lee_2017_unified}) to quantify the impact of each feature on the prediction. The model performed competitively in the closed track, surpassing the fine-tuned encoder baseline.

\paragraph{} Our analysis indicates two factors contributing to difficulty scores in the KVL data beyond the productive difficulty of words: the spelling difficulty and the choice of L1 equivalents and context in some test items.

\section{Related Research}

The task of vocabulary difficulty prediction, as construed by the BEA 2026 Shared Task \citep{bea2026st-findings}, effectively combines lexical complexity prediction (LCP) 
or complex word identification (CWI) 
with test item difficulty estimation. 

CWI is the task of identifying complex words in a sentence context.
CWI shared tasks \cite{paetzold-specia-2016-semeval,yimam-etal-2018-report} were dominated by feature-based systems focusing on word-level features.

LCP is an extension of CWI, where complexity is predicted on a continuous scale. The best performing approaches in an early shared task \citep{shardlow-etal-2021-semeval} were based on fine-tuning masked language models (MLMs). In the BEA 2024 Shared Task on the Multilingual Lexical Simplification Pipeline (MLSP; \citealp{shardlow-etal-2024-bea}), where lexical complexity was predicted for ten languages with limited training data, the top-performing systems were based either exclusively on LLM prompting or on word-level features. 

\citet{nohejl_etal_2025_japanese} proposed an LLM prompting-based LCP method, \textsc{G-Scale}, which aligns with the desired scale by applying temperature scaling to the probabilities and a linear regression to the final output. They also demonstrated that the addition of a single feature, log-frequency, can further improve the LLM-based predictions. 

\citet{smadu-etal-2024-investigating} compared feature-based models and fine-tuned Transformers (MLMs and LLMs) on CWI and LCP datasets with training data consisting of thousands of examples, i.e., in a setting similar to the present shared task. The study concluded that the LLMs rarely outperform the computationally less demanding MLMs and feature-based models. \citet{smadu-etal-2024-investigating} fine-tuned LLMs only using the standard cross-entropy loss with discretized complexity values as hard targets. We achieve large improvements in LLM and MLM fine-tuning performance by using soft targets to sidestep discretization.

Cross-entropy loss with soft targets is often associated with knowledge distillation \citep{hinton-etal-2015-distilling} and has been used for distilling MLMs and LLMs (e.g., \citealp{sanh-etal-2019-distilbert}). To the best of our knowledge, however, it has not been used for fine-tuning to predict continuous values.




Similar to LCP and CWI, the goal of the vocabulary difficulty prediction task is to estimate the difficulty of words given a specific context that determines their sense. There are, however, prominent differences:

\begin{enumerate}
\item The complexity in LCP and CWI is measured by \emph{subjective ratings}. The difficulty in the present task is based on the \emph{success rate of test items}.

\item The input for LCP and CWI is only a word and its context, both in the same language. The input for this task consists of multiple elements in English and in L1.

\item LCP and CWI focus on \emph{reading comprehension}. The present task focuses on \emph{written production}. 
\end{enumerate}

We address the specific aspects of this task while drawing on insights from prior research on LCP and CWI.

\citet{skidmore-etal-2025-transformer} fine-tuned encoders (i.e., MLMs) on the KVL data. Their fine-tuned XLM-RoBERTa model serves as the baseline of the shared task. They used SHAP as a tool for error analysis, attributing inaccurate predictions to specific input token positions. We use SHAP for explainability, attributing predictions to higher-level features such as spelling difficulty or similarity to L2.

\section{Method}

We propose two core methods: fine-tuning LLMs and MLMs with soft targets to predict the continuous difficulty values, and an explainable model with LLM-extracted features. To further improve the accuracy of both methods, we experiment with ensembling and additional features.

\subsection{Fine-Tuning with Soft Targets}\label{sec:method-llm}

In the following, we assume an LLM or an MLM that has a token vocabulary $V$ and predicts a probability distribution of tokens $i \in V$ conditioned on an input $\textbf{x}$, denoted as $\hat{p}(i \mid \mathbf{x})$.

The standard loss function for MLM and LLM training is a cross-entropy loss with a hard target, where the entire probability mass is assigned to a single target token, 
i.e., the negative log likelihood of the target token. Adapting such models to predict continuous values calls for a different approach.

While encoder models (typically MLMs) are sometimes fine-tuned for the prediction of continuous values using a regression head and mean squared error (MSE) loss, this is not the case for LLMs. The prevalent supervised fine-tuning (SFT) paradigm for LLMs is to convert such tasks to text generation by discretizing the continuous values and using the aforementioned cross-entropy with hard targets
\begin{equation}
\ell = -\log \hat p(v(d(y)) \mid \mathbf{x}),
\end{equation}
where $d$ is the discretization (e.g., rounding to the nearest integer), and $v$ is the mapping of discretized values to tokens.

These common approaches to predicting continuous values using MLMs and LLMs are reflected in the LCP methods based on encoders (e.g., \citealp{ide-etal-2023-japanese}), in the present shared task's encoder baseline \citep{skidmore-etal-2025-transformer}, and in the LCP methods based on LLMs (e.g., \citealp{smadu-etal-2024-investigating}). In contrast to the MSE loss used by the encoder-based regression, the standard cross-entropy loss for LLMs requires the continuous target values to be discretized into a small set of labels, losing precision in the process. 

Our method sidesteps this apparent misalignment between text generation and the prediction of continuous values by fine-tuning LLMs using cross-entropy loss with soft targets
\begin{equation}
\ell = - \sum_{i \in V} p(i) \log \hat p(i \mid \mathbf{x}).\label{eq:soft-ce}
\end{equation}
We prompt the model to predict values on a discrete scale $S$ in the form of successive integer points, e.g., $S = \{1, 2, 3, 4, 5\}$. Our aim, however, is to predict continuous values $y \in [\min S, \max S]$. We therefore express $y$ as a probability-weighted sum of its nearest points. Namely, we select two points, $a$ and $a + 1$, on the scale $S$ such that $a \le y \le a + 1$, and define the probability as
\begin{equation}
    p(i) = \begin{cases}
    (a+1)-y & \text{if } i = v(a), \\
    y-a     & \text{if } i = v(a+1), \\
    0       & \text{otherwise.}
    \end{cases}\label{eq:soft}
\end{equation}
We use $p$ as the soft target probability in the loss function defined in \autoref{eq:soft-ce}.

At inference time, we predict the token probability distribution $\hat{p}$ of a single token and then compute the final output $\hat{y}$ as a token probability-weighted mean: 
\begin{equation}
    \hat y = \frac{
        \sum_{s \in S} \hat{p} ( v(s) \mid \mathbf{x} ) \cdot s
        }{
        \sum_{s \in S} \hat{p} ( v(s) \mid \mathbf{x} )    
        }.\label{eq:prob-weighted}
\end{equation}
The same inference technique was used for LLMs by \citet{liu-etal-2023-g} to infer continuous scores using few-shot learning, and its variations have been applied to LCP \citep{aumiller-gertz-2022-unihd,enomoto-etal-2024-tmu,smadu-etal-2024-investigating,nohejl_etal_2025_japanese} but without the complementary loss function for fine-tuning. 

Because these two complementary techniques, soft-target cross-entropy loss and probability-weighted inference, require a single token to be predicted, they can be used not only for LLMs but also for MLMs via masked token prediction. For LLMs, the input $\mathbf{x}$ is simply the prompt; for MLMs, it is the sequence ``[CLS] \emph{prompt} [MASK] [SEP]''.


\subsection{Explainable Model}\label{sec:method-ex}
\label{sec:method-explainable}

Directly fine-tuning of LLMs or MLMs on this task is efficient but results in a black-box model, i.e., a model whose inner decision process is difficult to interpret. As an alternative model explainable using SHAP, we train an XGBoost \citep{chen_guestrin_2016_xgboost} regressor. SHAP provides an explanation of the model's output $f(\textbf{x})$ via SHAP values $\phi_i$ that additively express the local importance of each feature $x_i$:
\begin{equation}
f(\textbf{x}) = E[f(\textbf{x})] + \sum_{i=1}^{n} \phi_i.
\end{equation}
The SHAP values can be positive or negative, expressing how much each feature pushes a prediction higher or lower relative to the expected value. We use the additive nature of SHAP values and express the importance of groups of related features as sums of their SHAP values.

Our explainable model uses the following features:

\paragraph{Production frequency.} We use the log-frequency of the English word in the Lang-8 learner corpus \citep{mizumoto-etal-2011-mining}, to estimate its written production frequency by learners. Three features represent the subcorpora of (1) all learners, (2) L1 Spanish learners, and (3) L1 Chinese learners. There was not enough data for L1 German learners.

\paragraph{Reception frequency.} We use the log-frequency and log-range of the English word in corpora representative of spoken language to estimate its reception frequency. Namely, we use the logarithm of (1) frequency and (2) range (YouTube channels) in TUBE\-LEX \citep{nohejl-etal-2025-beyond}, and (3) frequency in the  spoken subcorpus of the British National Corpus (BNC; \citealpaliasyear{bncconsortium_2007_british}).

\paragraph{CEFR level.} We use the minimum CEFR level of the English word from the Cambridge English Vocabulary Profile \citep{capel_2012_completing}

\paragraph{Word length.} We measure the length of the English word in letters.

\paragraph{L1 similarity.} We compute the character-level similarity of the English word to the L1 word based on their length-normalized Levenshtein distance after removing diacritics and lowercasing. This feature is only applicable to languages written in the alphabet, Spanish and German.

\paragraph{Spelling difficulty.} We prompt GPT-5.2 \citep{openai_2025_update} to rate the spelling difficulty of an English word for each L1, given the L1 equivalent and the English word's pronunciation.

\paragraph{Lexical ambiguity.} We prompt (1) GPT-5.2 and (2) DeepSeek-V3.2 \citet{deepseek-ai_2025_deepseek} to determine the lexical ambiguity of the test item. A lexically ambiguous item must fulfill two conditions: (a) the English word is polysemous or one of several homonyms, and (b) the particular sense referenced by the test item is unfamiliar or challenging for learners.

\paragraph{L1 calque.} We prompt GPT-5.2 to determine if the English word is a morpheme-for-morpheme translation of the L1 equivalent. The prompt excludes single-morpheme words and simple borrowings but does not require one word to be a calque of the other in the etymological sense. In contrast to the L1 similarity feature, L1 calque is also applicable to Chinese, e.g., \ctext{超级英雄} `superhero' (composed of morphemes \ctext{超级} `super-' and \ctext{英雄} `hero').

\paragraph{}
For each of the prompt-based features, including the binary ones, we apply \textsc{G-Scale} \citep{nohejl_etal_2025_japanese}, i.e., temperature-scaled softmax followed by a probability-weighted mean of the LLM's predictions (\autoref{eq:prob-weighted}). The optimal temperature is determined for each model separately using cross-validation. This results in continuous feature values.

When reporting SHAP values, we sum the values within each group if it consists of multiple features (e.g., production frequency) without distinguishing the individual features.

We have submitted our explainable model as \texttt{explainable} and the model using only the traditional (not LLM-based) subset of features as \texttt{traditional}.

\subsection{Ensembles and Additional Features}\label{sec:method-ensemble-ft}

To further increase accuracy, we ensemble fine-tuned LLMs in a linear stack, combine fine-tuned LLM predictions with features, and experiment with additional features and fine-tuning encoder models using the same method we used for LLMs.

\paragraph{Linear stacking.} When fine-tuning LLMs, we perform 5-fold cross-validation and use the out-of-fold predictions to fit a linear regression that combines the predictions of multiple LLMs in an ensemble. The linear regression is fit separately for each L1. The final ensemble uses models fine-tuned on the complete data (the union of the provided test and development subsets). As this approach is rather computationally intensive, we compare it with simple average ensembling. Corresponding submission: \texttt{finetuned\_llms}.

\paragraph{Enhancing LLMs with features.} The linear stacking approach allows us to add our explainable features to the same linear regression as the fine-tuned LLMs, building on the \textsc{G-Scale} method proposed by \cite{nohejl_etal_2025_japanese}. Corresponding submission: \texttt{finetuned\_llms\_plus}.

\paragraph{Fine-tuned encoders and additional features.} We add an encoder model fine-tuned on single-language data and several other features to the feature model. We do this to maximize performance within the rules for the closed track, although the resulting model is no longer explainable. The added features are: (1) frequency in OpenSubtitles \citep{lison-etal-2018-opensubtitles2018}, a corpus similar to the already included TUBELEX; (2) frequency in the written subcorpus of the BNC; (3) CEFR level in the Global Scale of English\footnote{\url{https://www.english.com/gse/teacher-toolkit/user/vocabulary}} \citep{dejong_etal_2016_developing}, similar to the already included EVP level; (4) the Glasgow norms for concreteness, imageability, familiarity, and age of acquisition \citep{scott_etal_2019_glasgow}, which often strongly correlate with other already included features; and (5) an additional prompt for the calque feature. Corresponding submission: \texttt{closed\_max}. 

In the open track, we prompt two LLMs, GPT-4.1-mini and GPT-4.1-nano \citep{openai_2024_gpt}, to solve the test items, and we use their probability of a correct answer to indirectly assess the test item difficulty caused by the choice of the L1 word and context in the test item. To clearly distinguish this from the difficulty of the English  vocabulary itself, we call this feature \textbf{trickiness}. We hypothesize that the LLMs have near-perfect knowledge of common vocabulary in the four relevant languages, and their performance therefore reflects the trickiness of the test items rather than the difficulty of the English vocabulary. Corresponding submission: \texttt{finetuned\_llms\_plus}, \texttt{open\_max}.

We additionally experimented with prompting two recent LLMs, GPT-5.2 and GPT-4.1 \citep{openai_2024_gpt}, to directly rate the \textbf{difficulty} of the test items for learners of each L1 using 3-shot prompting. Corresponding submission: \texttt{open\_max}.

Neither of the last two prompting approaches (trickiness and difficulty) were permitted in the closed track.


\section{Experimental Settings}

We experimented with multiple prompts for fine-tuning using a small LLM and 1-epoch training on a single L1. However, as we show in the ablation analysis in \autoref{sec:abl}, with more training epochs, the prompt matters very little. For LLM prompting, we verified on a hand-picked sample that the model responses match our expectations, but we did not try to optimize the prompts.

In \autoref{app:prompts}, we provide complete listings of prompts used for fine-tuning and prompting.

As a basis for all fine-tuning experiments, we used pre-trained MLMs or LLMs, i.e., models not fine-tuned on chat or instruction data. For fine-tuning, we used recent LLMs with up to 32B parameters from the GLM-4 family \citep{teamglm_2024_chatglm}, Qwen2.5 family \citep{qwenteam_2025_qwen2}, and Ministral-3 family \citep{mistral_2026_ministral}, as well as MLMs \mbox{mmBERT} \citep{marone_etal_2025_mmbert} and XLM-RoBERTa \citep{conneau-etal-2020-unsupervised}. For prompting, we used models of the GPT-4.1 \citep{openai_2024_gpt} and GPT-5.2 \citep{openai_2025_update} families, and DeepSeek-V3.2 \citep{deepseek-ai_2025_deepseek}. \autoref{app:models} provides more details.

For MLMs, we performed end-to-end fine-tuning. For LLMs, we applied 4-bit quantization and fine-tuned QLoRA adapters \cite{dettmers_etal_2023_qlora} targeting all linear modules. All hyperparameters are described in \autoref{app:hyper}. We trained all models on the union of the test and development subsets provided by the shared task organizers. For the prompt-based features, we called models via the OpenAI API with zero temperature and requested log-probabilities necessary for probability weighting.

In all cases where a model is supposed to output a difficulty rating on a scale from 1 to 5, we map the difficulty scores from the KVL data linearly so that the highest score (the easiest item) in the training data maps to 5 and the lowest score to~1. The prompts are formulated accordingly. As the GLMM scores represent log-odds, we also experimented with mapping them to probabilities using the expit function (logistic curve), which decreased performance. See results in \autoref{app:probspace}.

\begin{table}[t]
\setlength{\tabcolsep}{3.1pt}
\footnotesize
\centering
\begin{tabularx}{\linewidth}{Xcccc}
\toprule
System & Chinese & German & Spanish & Mean \\
\midrule
\narrowtt{open\_max} & 0.631 & \textbf{0.723} & 0.743 & \textbf{0.699} \\
\narrowtt{finetuned\_llms\_plus} & \textbf{0.630} & 0.726 & \textbf{0.742} & 0.699 \\
\narrowtt{finetuned\_llms} & 0.640 & 0.731 & 0.760 & 0.710 \\
\midrule
$\le$32B LLM Average & 0.645 & 0.743 & 0.767 & 0.719 \\
- GLM-4-32B & 0.678 & 0.769 & 0.805 & 0.751 \\
- Qwen2.5-32B & 0.678 & 0.777 & 0.799 & 0.752 \\
- Ministral-3-14B & 0.681 & 0.781 & 0.799 & 0.753 \\
$\le$14B LLM Average & 0.662 & 0.770 & 0.804 & 0.745 \\
$\le$9B LLM Average & 0.683 & 0.791 & 0.835 & 0.769 \\
\midrule
Open-Track Baseline & 1.034 & 1.166 & 1.198 & 1.133 \\
Statistical Optimum & 0.321 & 0.304 & 0.205 & 0.277 \\
\bottomrule
\end{tabularx}
\caption{RMSE of our open-track submissions, compared with average ensembles by model size, individual models, and the shared task's open-track baseline.}
\label{tab:rmse-open-models-ensembles}
\end{table}

\section{Results}

In line with the shared task's evaluation, we report the root mean square error (RMSE) by L1 and compare it with the official shared tasks' baseline, a fine-tuned XLM-RoBERTa model \citep{bea2026st-findings}.

Note that RMSE is in the same units as the difficulty scores, which typically range from $-5$ to $+5$. 
We also report the Pearson's correlation coefficients (PCC) $r$, the secondary evaluation metric, which may be easier to interpret, in \autoref{app:pcc}.

To further put the results into perspective, we report a ``Statistical Optimum'' result, which simulates the largest error (lowest correlation) that should be considered optimal given the precision of the gold standard data. The simulation is based on confidence intervals reported by \citet{schmitt_etal_2024_knowledge}; see \autoref{app:stat-opt} for details. 

\subsection{Open Track}

As shown in \autoref{tab:rmse-open-models-ensembles}, the results of our three open-track submissions are very close to each other. In summary, the results do not justify the cost of adding features to fine-tuned LLMs. This contrasts with previous findings for in-context learning setting, where just adding frequency as a feature improves LLM predictions of lexical complexity \citep{nohejl_etal_2025_japanese}. The models with added features (\texttt{open\_max} and \texttt{finetuned\_llms\_plus}) surpassed all other submissions in the open track on all three languages. The linear stack of LLMs (\texttt{finetuned\_llms}) surpassed all other submissions on Chinese and German but was narrowly outperformed by submissions from the Glite team on Spanish. The RMSE of approximately 0.7 that our models achieve is relatively close to the statistical optimum of 0.277. 

Linear stacking (\texttt{finetuned\_llms}) results in a modest improvement over the average ensemble of the same models ($\le$32B LLM Average). The ensembling itself, however, improves performance more substantially, from an RMSE of 0.751--0.753 for individual models to 0.719 for the average ensemble. While increasing the model size obviously improves performance, it is worth noting that the smaller Ministral-3-14B model performs on par with the 32B Qwen2.5 and GLM-4 models.

In \autoref{app:model-size}, we compare all individual LLMs used in the ensembles above, as well as LLMs with 0.5B parameters and similarly sized MLMs, showing that LLMs and MLMs achieve comparable results at comparable model sizes.

\begin{table}[t]
\setlength{\tabcolsep}{3.1pt}
\footnotesize
\centering
\begin{tabularx}{\linewidth}{Xcccc}
\toprule
System & Chinese & German & Spanish & Mean \\
\midrule
\texttt{closed\_max} & \textbf{0.816} & \textbf{0.963} & \textbf{0.983} & \textbf{0.921} \\
\texttt{explainable} & 0.920 & 1.126 & 1.156 & 1.067 \\
\texttt{traditional} & 1.078 & 1.195 & 1.305 & 1.193 \\
\midrule
\texttt{expl}.\ -- std.\ inference & 0.961 & 1.151 & 1.190 & 1.101 \\
\texttt{expl}.\ -- lin.\ regression & 0.975 & 1.154 & 1.202 & 1.111 \\
\midrule
Closed-Track Baseline & 1.140 & 1.258 & 1.257 & 1.218 \\
\bottomrule
\end{tabularx}
\caption{RMSE of our closed-track submissions, compared with two variants of the \texttt{explainable} model and the shared task's closed-track baseline.}
\label{tab:rmse-closed-models}
\end{table}

\subsection{Closed Track}

\autoref{tab:rmse-closed-models} compares the results of our submissions to the closed-track baseline. Interestingly, our model with only traditional features performs on par with the encoder-based baseline, and our \texttt{explainable} model with extra LLM-based features outperforms the baseline by a relatively wide margin (mean RMSE of 1.218 vs.\ 1.067). The \texttt{closed\_max} model achieves a further improvement (mean RMSE of 0.921) by adding a fine-tuned MLM as a feature.

Two variants applied to the explainable features---standard inference for prompt-based features instead of probability weighting, and linear regression instead of an XGBoost regressor---result in higher RMSE.

\begin{table}[t]
\setlength{\tabcolsep}{3.1pt}
\footnotesize
\centering
\begin{tabularx}{\linewidth}{Xcccc}
\toprule
Method (Base Model) & Chinese & German & Spanish & Mean \\
\midrule
\makebox[0pt][l]{Ours (Ministral-3-14B)} & \textbf{0.681} & \textbf{0.781} & \textbf{0.799} & \textbf{0.753} \\
- single language & 0.701 & 0.824 & 0.799 & 0.775 \\
- out-of-language & 0.874 & 0.901 & 0.999 & 0.925 \\
- short prompt & 0.697 & 0.790 & 0.808 & 0.765 \\
- standard loss & 0.762 & 0.841 & 0.892 & 0.832 \\
- std.\ loss \& inference & 0.863 & 0.943 & 1.003 & 0.936 \\
\bottomrule
\end{tabularx}
\caption{RMSE of ablations of our LLM-based model.}
\label{tab:rmse-llm-ablation}
\end{table}

\begin{table}[t]
\setlength{\tabcolsep}{3.1pt}
\footnotesize
\centering
\begin{tabularx}{\linewidth}{Xcccc}
\toprule
Method (Base Model) & Chinese & German & Spanish & Mean \\
\midrule
Ours (mmBERT-b) & 0.921 & \textbf{0.984} & \textbf{1.063} & \textbf{0.989} \\
- single language & \textbf{0.916} & 1.044 & 1.070 & 1.010 \\
- out-of-language & 1.115 & 1.102 & 1.193 & 1.136 \\
- short prompt & 0.929 & 1.012 & 1.066 & 1.002 \\
- standard loss & 1.003 & 1.099 & 1.166 & 1.089 \\
- std.\ loss \& inference & 1.064 & 1.144 & 1.192 & 1.133 \\
\midrule
Regression (XLMR-b) & 1.222 & 1.276 & 1.373 & 1.290 \\
Regression (XLMR-l) & 1.048 & 1.144 & 1.192 & 1.128 \\
Reg.\ (mmBERT-b) & 1.000 & 1.105 & 1.153 & 1.086 \\
\bottomrule
\end{tabularx}
\caption{RMSE of ablations of our MLM-based model, compared with using a standard regression head and different base models (XLM-RoBERTa-base/large).}
\label{tab:rmse-mlm-ablation}
\end{table}

\section{Analysis and Discussion}

\subsection{Ablation Analysis}\label{sec:abl}

We ablate the fine-tuned LLM and MLM that were used in our open-track submissions and the \texttt{closed\_max} closed-track submission. For the open-track submission, we ablate only one of the three similarly performing LLMs in the ensemble, Ministral-3-14B as follows:

\paragraph{Single language.} We fine-tune the model on single L1 data instead of all three languages.

\paragraph{Out-of-language.} We fine-tune the model on data for the other two L1s (e.g.\ the model tested on Chinese is fine-tuned on German and Spanish).

\paragraph{Short prompt.} We use a minimalistic prompt template instead of the basic verbose one. See \autoref{app:ft-prompts}.

\paragraph{Standard loss.} Instead of using the cross-entropy loss with a soft target, described in \autoref{sec:method-llm}, we fine-tune using the standard cross-entropy loss with the discretized score following \citet{smadu-etal-2024-investigating}.

\paragraph{Standard loss and inference.} We fine-tune using the standard cross-entropy loss with a discretized score, and instead of using probability-weighting at inference, we decode the highest probability token.

In addition to performing the same ablations for the MLM, we also compare the results of our method to those of using a regression head with MSE loss on mmBERT-base and two sizes of XLM-RoBERTa models (listed as \textbf{Regression}). Note that in the closed track, rules required us to use the variant we report here as an ablation (mmBERT-b/single language).

As shown in \autoref{tab:rmse-llm-ablation} and \autoref{tab:rmse-mlm-ablation}, our method outperforms all ablations with the single exception of the MLM model fine-tuned only on Chinese L1 data, which performs better on Chinese by a small margin. In the other cases, training on a single language results in only a small performance drop.

The standard loss ablation demonstrates that our technique of fine-tuning using cross-entropy loss with soft targets is superior to common approaches for both LLMs and MLMs. Fine-tuning using the standard loss results in a tangible performance drop for both model types (a difference in mean RMSE of 0.079 and 0.900, respectively). For the MLM, the common approach of fine-tuning a regression head with MSE loss degrades performance similarly (by a difference in mean RMSE of 0.809).

\subsection{The Explainable Models}

As outlined in \autoref{sec:method-explainable}, we designed a feature-based model with traditional and LLM-based features that all have a clear interpretation. SHAP values explain the model's output locally in terms of the impact of each feature, which has both sign and magnitude. For a global analysis of feature importance, we report the mean absolute SHAP values computed over the test data in \autoref{fig:global-shap}.

\newcommand{\shifttitle}{\hspace{0.21\linewidth}}
\newcommand{\niceskip}{\vspace{6pt}}

\begin{figure}[t]
    \sffamily
    \smaller[2]
    \centering
    \shifttitle Chinese\par
    \includegraphics[width=1.0\linewidth]{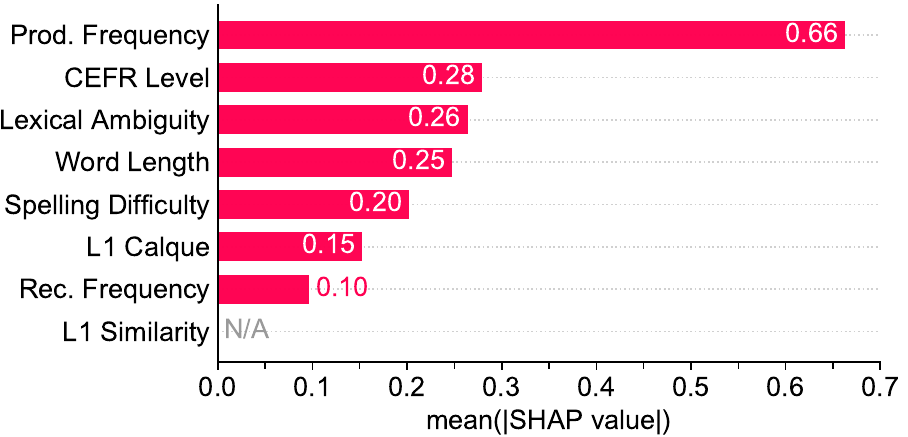}\par\niceskip
    \shifttitle German\par
    \includegraphics[width=1.0\linewidth]{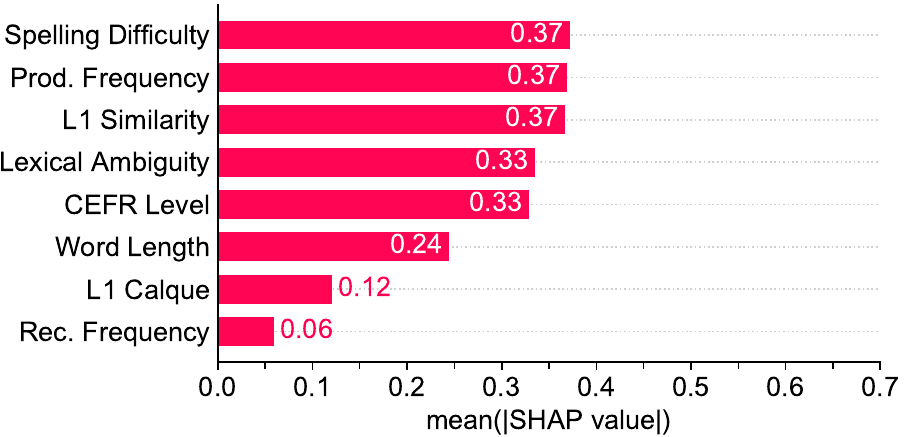}\par\niceskip
    \shifttitle Spanish\par
    \includegraphics[width=1.0\linewidth]{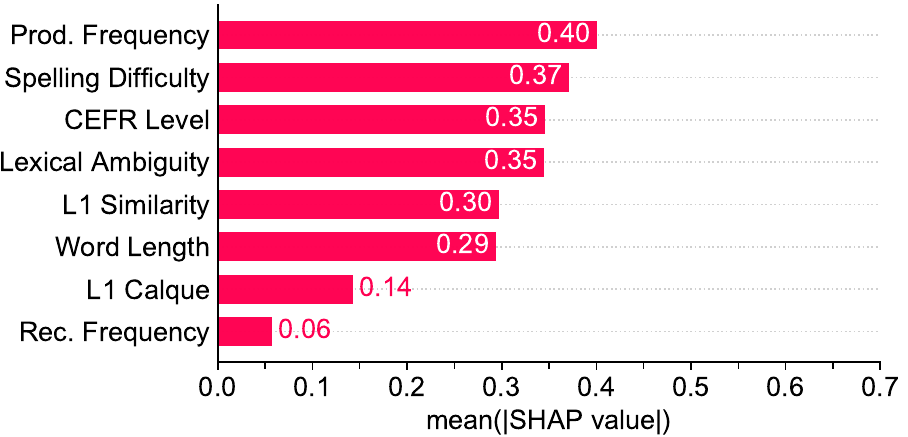}
    \caption{Global SHAP summaries by L1.}
    \label{fig:global-shap}
\end{figure}

\renewcommand{\niceskip}{\vspace{5.2pt}}

\begin{figure}[t]
    \sffamily
    \smaller[2]
    \hfill Chinese: \smash{\ctext{农庄住宅（农场主的房子）}\hspace{-0.5em}, Prediction: -0.20, Target: 0.14}\par

\centering    \includegraphics[width=1.0\linewidth]{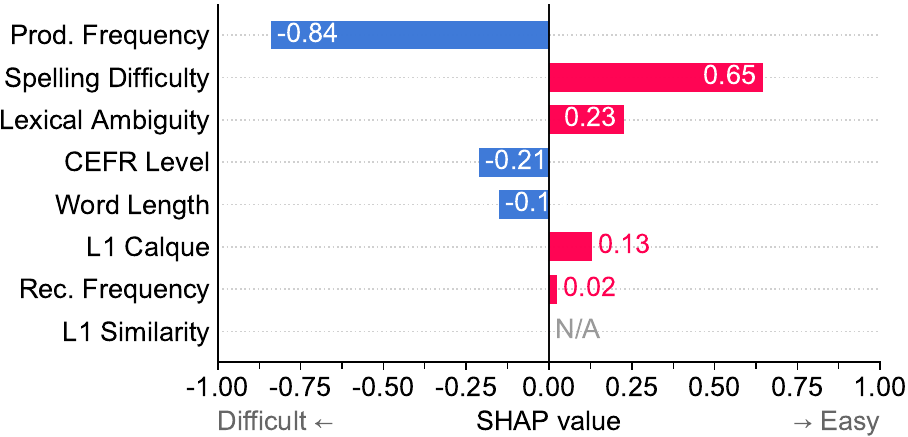}\par\niceskip
    \shifttitle German: Bauernhaus, Prediction:s 0.65, Target: 0.94\par
    \includegraphics[width=1.0\linewidth]{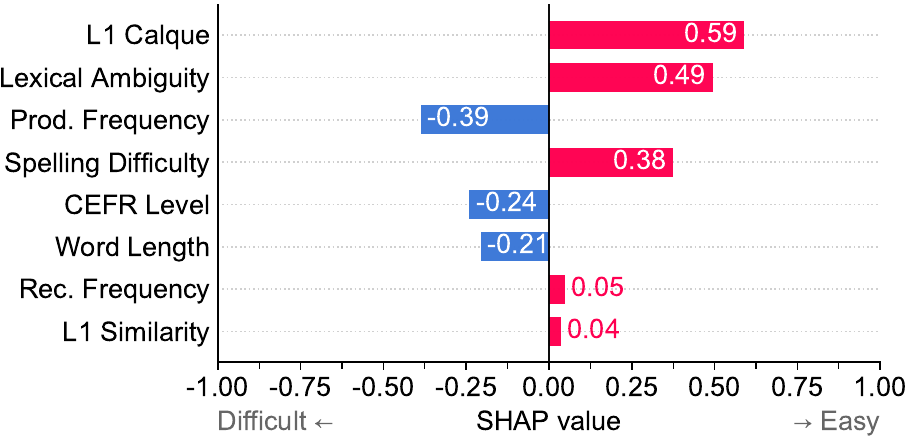}\par\niceskip
    \shifttitle Spanish: alquería, Prediction: -1.87, Target: -2.14\par
    \includegraphics[width=1.0\linewidth]{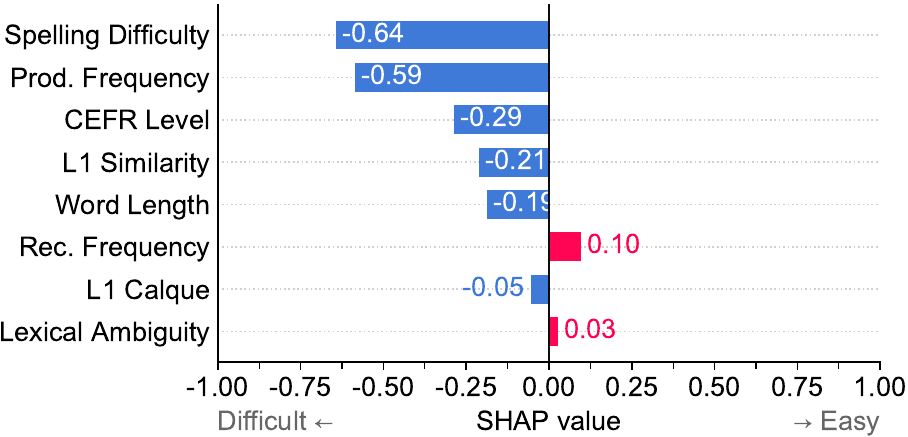}
    \caption{Example of local SHAP explanations by L1 for the English word \textit{farmhouse}.}
    \label{fig:local-shap}
\end{figure}

\textbf{Production frequency} is a consistently high-impact feature across languages. Its importance per language, in fact, corresponds with the amount of L1-specific data available. Yet, even for German, with no L1-specific data available, the feature almost completely overshadows \textbf{reception frequency}, which ranks as the least important across L1s. This is in line with the focus of the KVL data on production and can be contrasted with findings for the comprehension-focused LCP, where reception frequency (represented by frequency in TUBELEX) and production frequency (represented by frequency in Lang-8) are equally good predictors \citep{nohejl-etal-2024-difficult}

\textbf{Spelling difficulty} is the most important feature for L1 German and the second most important feature for Spanish. Perhaps counter-intuitively, it is much less important for L1 Chinese. We hypothesize this is caused by two factors. First, the production frequency uses learner-written texts, and therefore it partially discounts the frequency of words with frequent mistakes. As a result, the importance of the separate spelling difficulty feature is lower proportionally to the size of the L1-specific written production data. Second, for Chinese speakers, there may be less interference from the orthography of their L1, which is primarily written in Chinese characters, while Spanish and German speakers may be more prone to making errors due to the influence of their L1's spelling (e.g., English \textit{music}, \textit{action}; German \textit{Musik}, \textit{Aktion}).

\textbf{CEFR level} ranks particularly high for Chinese L1 learners. This corresponds to a very small proportion of cognates between English and Chinese compared to the two European languages, where \textbf{L1 Similarity} can play a comparable (Spanish) or more important role (German).

\textbf{Lexical ambiguity} has a similarly high impact as the CEFR level and can be seen as complementary to the CEFR level and the frequency features, all of which focus on the surface lemma (frequency of all senses or homonyms, the lowest CEFR level, regardless of sense). We recognize words as lexically ambiguous if the test item refers to a less common sense or a homonym.

\textbf{Word length} is surprisingly one of the consistently relevant features. While length certainly contributes to the difficulty in acquiring a word, it also simply increases the probability of typos, even in cases when the writer knows the word's spelling.

\textbf{L1 calque} is one of the globally least important features, reflecting both the frequency of calques in the language pairs and the fact that an L1 word being a calque (morpheme-for-morpheme translation) being less conducive to eliciting the production of the correct English word than the surface level L1 similarity.

Local explanations of all test set predictions of the explainable model can be viewed in an online application.\footnote{\url{https://ynklab.github.io/vocabulary-difficulty/}} In \autoref{fig:local-shap} we provide an example of the local explanation of the difficulty of the English word \textit{farmhouse}, which, despite being relatively rare and not being assigned any CEFR level in the EVP, is not a particularly difficult word for L1 Chinese and German learners. Both its Chinese and German counterparts are calques of the English word \textit{farmhouse}, which contributes to lower difficulty in these languages. On the other hand, for L1 Spanish speakers, the word is assessed as comparatively difficult to spell.


Neither the global nor the local explanations can be interpreted as the actual causes of the difficulty of a particular word. In particular, it cannot tell us anything about features not present in the model, and it underestimates the importance of a factor when the data has low quality (e.g., production frequency in the case of L1 German). On the other hand, given the relatively high accuracy of the model, we believe that the globally high-impact features provide valuable insights.

Both the high impact of production frequency and spelling difficulty is in line with the focus of the KVL. However, the combined importance of features reflecting the difficulty of spelling a word without errors (spelling difficulty, word length, and, to some extent, also the learners' written production frequency) could affect the utility of the dataset for other uses, such as creating reading materials matched to the learner's level. \cite{schmitt_etal_2024_knowledge} suggest the KVL data could be more suitable for this purpose than frequency.

Given that the KVL are designed to focus on the most common word sense \cite{schmitt_etal_2024_knowledge}, the high importance of lexical ambiguity in our analysis could seem spurious. We have, however, observed uncommon word senses in the data. For instance, the noun \textit{wireless} is included in the sense `radio (receiving set)', which is dated and mainly British according to the Oxford Dictionary of English, rather than in the sense `wireless broadcasting or networking using radio signals'. For other words, such as \textit{log} (`chunk of wood' in KVL, not `record') or \textit{diet} (`nutrition' in KVL, not `dietary regime'), two senses may be similarly common. Lexical ambiguity provides a useful signal in all these cases.

\subsection{Tricky Test Items}

The shared task rules permitted us to use the trickiness feature only in the open-track submission, where its impact was relatively small.  However, it provides an interesting insight into how the test item construction often affects the resulting difficulty scores. The value of our trickiness feature is the probability of an LLM answering the test item incorrectly. 
\begin{table}[t]
    \centering\footnotesize
    \begin{tabular}{lll}
    \toprule
        English word & Spanish word & Top LLM response\\
    \midrule
        \textbf{i}nstantly &	inmediatamente & \textbf{i}mmediately*\\
        \textbf{m}otivational & estimulante & \textbf{m}otivating*\\
        (to) \textbf{r}eform	& enmendar & \textbf{r}evise\\
        (to) \textbf{a}mount	& sumar & \textbf{a}dd up\\
        \textbf{e}verybody &	todo el mundo &	\textbf{e}veryone*\\
        \textbf{s}ynonymous & sinónimo & \textbf{s}ynonym*\\
\bottomrule
    \end{tabular}
    \caption{Examples of tricky test items in L1 Spanish, based on responses of GPT-4.1-mini. First letter printed in bold. Asterisks mark incorrect word length.}
    \label{tab:tricky}
\end{table}
\autoref{tab:tricky} shows examples of high-trickiness items from the test set. For brevity, we focus on cases that do not depend on the L1 context. As exemplified by the items \textit{instantly}, \textit{motivational}, and \textit{reform}, in many cases, the creators of the KVL test items intentionally avoided cognate L1 words. The words they have chosen instead often correspond better to English words other than the intended ones. In other cases, the reason is not the avoidance of cognates, but the L1 word strongly suggests a different English word nonetheless. While in some cases the L1 context or the number of letters disambiguates the English word (e.g., \textit{sinónimo} is used as an adjective, not a noun; \textit{everybody} and \textit{everyone} differ in word length), such details were likely easy to miss not only for the LLM but also for the respondents. We believe that while better choices could have been made for some items, this shows an inherent limitation of the test item format used to compile the KVL.



\section{Conclusion}

We proposed a novel method for fine-tuning LLMs and MLMs to predict continuous values using cross-entropy loss with soft targets. The method achieved consistent gains on the vocabulary difficulty prediction task over prior approaches, such as discretization for LLMs and a regression head with MSE loss for MLMs.

In our fine-tuning experiments, increases in base model size up to 32B parameters resulted in performance improvements. The larger-sized LLMs also consistently outperformed the smaller-sized MLMs. This result differs from the LCP study by \citet{smadu-etal-2024-investigating}, where much smaller fine-tuned MLMs performed comparably to much larger LLMs. We hypothesize that this could be due to the more complex nature of vocabulary difficulty prediction compared to LCP. 

We built a separate explainable model that achieves competitive results on the task and provides insights into what affects vocabulary difficulty scores in the British Council's KVL data. The high impact of spelling difficulty and the construction of the test items call for further investigation.

\section*{Limitations}

We tested the proposed method for fine-tuning LLMs and MLMs to predict continuous values using cross-entropy loss with soft-targets only on this shared task's data (KVL). Its suitability for other tasks and settings is yet to be investigated. On smaller or noisier data, for instance, its impact may be less pronounced. Various parameters, such as the number of points on the scale used, may also affect its performance.

The insights provided by our explainable model and the trickiness model can, in principle, be empirically validated, but we have not been able to do so, as the individual responses used to compile KVL are not publicly available. The frequency of spelling mistakes would be particularly easy to validate with access to the responses.


\bibliography{anthology-1,anthology-2,bea2026st-sakura,bea2026st-findings}

\appendix

\section{Prompt Templates}
\label{app:prompts}

In the following, we list the full text of our prompt templates by feature for which they were used. The symbol $\dlsh$ stands for line breaks. Placeholders are printed in bold. The following common placeholders corresponding to the dataset items are used across multiple prompts:

\begin{itemize}
  \item \texttt{\{l1\_name\}}: an L1, i.e., \texttt{Spanish}, \texttt{Chinese}, or \texttt{German}.
  \item \texttt{\{l1\_word\}}: an L1 word.
  \item \texttt{\{l1\_context\}}: an L1 context.
  \item \texttt{\{en\_word\}}: an English word.
  \item \texttt{\{clue\}}: a letter-pattern clue derived from the English word (first letter + space-separated underscores), e.g., \texttt{b \_ \_ \_} for \texttt{book}.
  \end{itemize}
  
\subsection{Fine-Tuning}
\label{app:ft-prompts}

\subsubsection*{Basic template:}

\begin{raggedright}%
\noindent%
\ttfamily\footnotesize\microtypecontext{protrusion=false}%
Rate how difficult it is for learners to guess the English word based on the \textbf{\{l1\_name\}} word, context and clue on a scale from 1 to 5 (1=very easy, 5=very difficult).\ttnl
\textbf{\{l1\_name\}} word: \textbf{\{l1\_word\}}\ttnl
\textbf{\{l1\_name\}} context: \textbf{\{l1\_context\}}\ttnl
Clue: \textbf{\{clue\}}\ttnl
English word: \textbf{\{en\_word\}}\ttnl
Difficulty:\\
\end{raggedright}

\subsubsection*{Short prompt template:}
Used in the ``short prompt'' ablation.
\smallskip

\begin{raggedright}%
\noindent%
\ttfamily\footnotesize\microtypecontext{protrusion=false}%
\textbf{\{l1\_word\}} \#\#\# \textbf{\{l1\_context\}} \#\#\# \textbf{\{clue\}} \#\#\# \textbf{\{en\_word\}} \#\#\# Difficulty (1 to 5):
\end{raggedright}

\subsubsection*{Regression template}

Used for MLMs when fine-tuning with a regression head, following \citet{skidmore-etal-2025-transformer}.
\smallskip

\begin{raggedright}%
\noindent%
\ttfamily\footnotesize\microtypecontext{protrusion=false}%
[CLS] \textbf{\{l1\_word\}} [SEP] \textbf{\{l1\_context\}} [SEP] \textbf{\{clue\}} [SEP] \textbf{\{en\_word\}} [SEP]
\end{raggedright}

\subsection{Lexical Ambiguity}

\subsubsection*{Placeholders:}

We use the English word ``bank'' as an example of lexical ambiguity (`financial institution' vs.\ `(river) side'). For L1 Spanish, the placeholders would take the following values:

\begin{itemize}
  \item \texttt{\{ex\_en\_word\}}: ``bank''
  \item \texttt{\{ex\_easy\_word\_l1\}}: ``banco''
  \item \texttt{\{ex\_easy\_context\_l1\}}: ``Deposité el dine\-ro en el banco.''
  \item \texttt{\{ex\_hard\_word\_l1\}}: ``orilla''
  \item \texttt{\{ex\_hard\_context\_l1\}}: ``Nos sentamos en la orilla del río.''
\end{itemize}

\subsubsection*{Prompt template:}

\begin{raggedright}%
\noindent%
\ttfamily\footnotesize\microtypecontext{protrusion=false}%
You are a language education expert.\ttnl

TASK\ttnl
Given:\ttnl
- an English word form (the "English word"),\ttnl
- an L1 gloss/translation (the "\textbf{\{l1\_name\}} item"),\ttnl
- and the L1 usage context sentence (the "\textbf{\{l1\_name\}} context"),\ttnl
decide whether the English word, when used to express the meaning suggested by the L1 item + context,\ttnl
meets BOTH conditions:\ttnl

A) Lexical ambiguity: the English word has multiple established senses that share the same form\ttnl
\ \ \ (polysemy or homonymy), such that another common sense could plausibly be activated/confused.\ttnl

B) Unfamiliarity for L2 learners: in this meaning/usage, the English word is likely to be unfamiliar\ttnl
\ \ \ or challenging for typical second-language learners (e.g., less frequent sense, idiomatic/figurative,\ttnl
\ \ \ domain-specific usage, nonliteral extension).\ttnl

OUTPUT REQUIREMENTS\ttnl
- Output "1" if BOTH conditions (A and B) are met; otherwise output "0".\ttnl
- Output MUST be exactly one character: 1 or 0.\ttnl
- Do NOT include explanations, alternatives, quotes, or extra text.\ttnl

EXAMPLE 1\ttnl
English word: \textbf{\{ex\_en\_word\}}\ttnl
\textbf{\{l1\_name\}} item: \textbf{\{ex\_easy\_word\_l1\}}\ttnl
\textbf{\{l1\_name\}} context: \textbf{\{ex\_easy\_context\_l1\}}\ttnl
Is the English word ambiguous and unfamiliar: 0\ttnl

EXAMPLE 2\ttnl
English word: \textbf{\{ex\_en\_word\}}\ttnl
\textbf{\{l1\_name\}} item: \textbf{\{ex\_hard\_word\_l1\}}\ttnl
\textbf{\{l1\_name\}} context: \textbf{\{ex\_hard\_context\_l1\}}\ttnl
Is the English word ambiguous and unfamiliar: 1\ttnl

NOW DECIDE\ttnl
English word: \textbf{\{en\_word\}}\ttnl
\textbf{\{l1\_name\}} item: \textbf{\{l1\_word\}}\ttnl
\textbf{\{l1\_name\}} context: \textbf{\{l1\_context\}}\ttnl
Is the English word ambiguous and unfamiliar:\\
\end{raggedright}

\subsection{Spelling Difficulty}

This prompt is executed for all L1s at once.

\subsubsection*{Placeholders:}
\begin{raggedright}
\begin{itemize}
  \item \texttt{\{en\_pron\}}: CMU-style pronunciation string of the English word of the current item.
  Example: \texttt{K Y UW} for ``queue''.
  \item \texttt{\{all\_l1\_words[l1]\}}: L1 words of the current item (all three in one prompt).
 
 \item \texttt{\{hard\_pron\}}, \texttt{\{hard\_cn\}}, \texttt{\{hard\_es\}}, \texttt{\{hard\_de\}}, \texttt{\{hard\_cn\_score\}}, \texttt{\{hard\_es\_score\}}, \texttt{\{hard\_de\_score\}}:
  hard demonstration item (pronunciation, three L1 translations, and example scores 5, 4, 4).

  \item \texttt{\{easy\_pron\}}, \texttt{\{easy\_cn\}}, \texttt{\{easy\_es\}}, \texttt{\{easy\_de\}}, \texttt{\{easy\_cn\_score\}}, \texttt{\{easy\_es\_score\}}, \texttt{\{easy\_de\_score\}}:
  easy demonstration item (pronunciation, three L1 translations, and example scores 1, 1, 1).

\end{itemize}
\end{raggedright}

\subsubsection*{Prompt template:}
\begin{raggedright}%
\noindent%
\ttfamily\footnotesize\microtypecontext{protrusion=false}%
TASK\ttnl
You are required to rate English spelling difficulty on a 1--5 scale, where 1 = very easy and 5 = very difficult.\ttnl
You will be given English pronunciation and the target word's translation in Chinese, Spanish, and German.\ttnl
Evaluate how difficult it would be for learners with Chinese, Spanish, and German L1 backgrounds to spell the English word with that pronunciation correctly when they know the translation in their native language.\ttnl

OUTPUT REQUIREMENTS\ttnl
- Output exactly one digit (1, 2, 3, 4, or 5) for each L1, separated by commas, in the order of Chinese, Spanish, German.\ttnl
- Do not include any other text.\ttnl

EXAMPLE 1\ttnl
English pronunciation: '\textbf{\{hard\_pron\}}'\ttnl
Chinese: \textbf{\{hard\_cn\}}\ttnl
Spanish: \textbf{\{hard\_es\}}\ttnl
German: \textbf{\{hard\_de\}}\ttnl
Result: \textbf{\{hard\_cn\_score\}},\textbf{\{hard\_es\_score\}},\textbf{\{hard\_de\_score\}}\ttnl

EXAMPLE 2\ttnl
English pronunciation: '\textbf{\{easy\_pron\}}'\ttnl
Chinese: \textbf{\{easy\_cn\}}\ttnl
Spanish: \textbf{\{easy\_es\}}\ttnl
German: \textbf{\{easy\_de\}}\ttnl
Result: \textbf{\{easy\_cn\_score\}},\textbf{\{easy\_es\_score\}},\textbf{\{easy\_de\_score\}}\ttnl

NOW DECIDE\ttnl
English pronunciation: \textbf{\{en\_pron\}}\ttnl
Chinese: \textbf{\{all\_l1\_words['cn']\}}\ttnl
Spanish: \textbf{\{all\_l1\_words['es']\}}\ttnl
German: \textbf{\{all\_l1\_words['de']\}}\ttnl
Result:\\
\end{raggedright}

\subsection{L1 Calque}

This prompt uses Spanish examples for all L1s.

\subsubsection*{Prompt template:}
\begin{raggedright}%
\noindent%
\ttfamily\footnotesize\microtypecontext{protrusion=false}%
You are a linguist and your task is to decide whether an English word is a morpheme-for-morpheme translation of any of the given \textbf{\{l1\_name\}} equivalents.\ttnl
The morpheme-for-morpheme mapping must be 1:1. 1:N or other mappings do not count.\ttnl
Single morpheme translations or simple borrowings/cognates do not count either.\ttnl
Respond only with YES or NO.\ttnl

wave/ola: NO (reason: single morpheme)\ttnl
ecosystem/ecosistema: NO (reason: simple cognate)\ttnl
hotdog/perro caliente: YES (reason: hot=caliente, dog=perro)\ttnl
stare/mirar fijamente: NO (reason: not a 1:1 mapping)\ttnl
\textbf{\{en\_word\}}/\textbf{\{l1\_word\}}:\\
\end{raggedright}

\subsection{L1 Calque (used only in \texttt{closed\_max})}

This was our first iteration of the prompt. It did not give the expected results (monomorphemic word pairs were labeled as calques), so we did not use it for our explainable model. However, as it performed well as a feature, we included it in the \texttt{closed\_max} model.

\subsubsection*{Placeholders:}
\begin{itemize}
  \item \texttt{\{ex\_calque\_l1\}}: the L1-side word, e.g., ``\ctext{热狗}'' for Chinese, composed of the morphemes `hot' and `dog'.
  \item \texttt{\{ex\_calque\_en\}}: the English-side word, e.g., ``hot dog''.
\end{itemize}

\subsubsection*{Prompt template:}
\begin{raggedright}%
\noindent%
\ttfamily\footnotesize\microtypecontext{protrusion=false}%
You are a bilinguistics expert.\ttnl

TASK\ttnl
Given a \textbf{\{l1\_name\}} item and an English item, decide whether there exists a best-matching candidate in the \textbf{\{l1\_name\}} item that is a component-by-component (morpheme-level) translation of the English item.\ttnl

A component-by-component mapping means that the meaningful parts\ttnl
(words, roots, prefixes, or suffixes) of the English item are directly translated\ttnl
into corresponding meaningful parts in the \textbf{\{l1\_name\}} item.\ttnl

Procedure (internal; do NOT output these steps):\ttnl
1) If the \textbf{\{l1\_name\}} item contains multiple candidates, select exactly ONE candidate: the one that aligns best component-wise with the English form.\ttnl
2) Judge ONLY that selected candidate for component-by-component mapping.\ttnl

OUTPUT REQUIREMENTS\ttnl
- Output "1" if the selected best candidate is a component-by-component mapping; otherwise output "0".\ttnl
- Output MUST be exactly one character: 1 or 0.\ttnl
- Do NOT include explanations, alternatives, quotes, or extra text.\ttnl

EXAMPLE\ttnl
\textbf{\{l1\_name\}} item: \textbf{\{ex\_calque\_l1\}}\ttnl
English item: \textbf{\{ex\_calque\_en\}}\ttnl
Is word-for-word mapping: 1\ttnl

NOW DECIDE\ttnl
\textbf{\{l1\_name\}} item: \textbf{\{l1\_word\}}\ttnl
English item: \textbf{\{en\_word\}}\ttnl
Is word-for-word mapping:\\
\end{raggedright}

\subsection{Trickiness (short prompt, used only in \texttt{open\_max})}\label{app:short-solve}

\subsubsection*{Placeholders:}
\texttt{\{solve\_example\}}: a formatted one-shot example (L1 word, L1 context, English word) using the item English word ``strawberry'. Example:
\medskip

\begin{raggedright}%
\noindent%
\ttfamily\footnotesize\microtypecontext{protrusion=false}%
German word: Erdbeere\ttnl{}
German context: Ich mag keine Erdbeeren.\ttnl
English word: strawberry\\
\end{raggedright}

\subsubsection*{Prompt template:}
\begin{raggedright}%
\noindent%
\ttfamily\footnotesize\microtypecontext{protrusion=false}%
You are bilingual in \textbf{\{l1\_name\}} and English and your task is to find the best English translation for a \textbf{\{l1\_name\}} word given a context and constraints. The constraints are given in the form of a clue, e.g., "b \_ \_ \_", meaning that the word starts with the (upper or lower case) letter B and has 4 letters. You must give a single English word in dictionary form (lemma) as a response.\ttnl

\textbf{\{solve\_example\}}\ttnl
\textbf{\{l1\_name\}} word: \textbf{\{l1\_word\}}\ttnl
\textbf{\{l1\_name\}} context: \textbf{\{l1\_context\}}\ttnl
Clue: \textbf{\{clue\}}\ttnl
English word:\\
\end{raggedright}

\subsection{Trickiness (long prompt, used only in \texttt{open\_max})}

\subsubsection*{Placeholders:}
Same as in \autoref{app:short-solve}.

\subsubsection*{Prompt template:}
\begin{raggedright}%
\noindent%
\ttfamily\footnotesize\microtypecontext{protrusion=false}%
You are bilingual in \textbf{\{l1\_name\}} and English.\ttnl

TASK\ttnl
Given a word in \textbf{\{l1\_name\}}, its usage context, and a spelling clue, find the single best English translation that fits BOTH the meaning and the spelling constraint.\ttnl

INPUTS\ttnl
- \textbf{\{l1\_name\}} word: a single word to translate\ttnl
- \textbf{\{l1\_name\}} context: a sentence showing how the word is used\ttnl
- Clue: a pattern such as "b \_ \_ \_", where:\ttnl
\ \ * the first letter is indicated (case-insensitive)\ttnl
\ \ * "\_" indicates subsequent unknown letter\ttnl
\ \ * the total number of letters must match exactly\ttnl

OUTPUT REQUIREMENTS\ttnl
- Output EXACTLY ONE English word\ttnl
- The word must be:\ttnl
\ \ * a dictionary form (lemma)\ttnl
\ \ * a single token (no spaces, hyphens, or punctuation)\ttnl
\ \ * consistent with the context\ttnl
\ \ * consistent with the clue\ttnl
- Do NOT include explanations, alternatives, quotes, or extra text.\ttnl

EXAMPLES\ttnl
\textbf{\{solve\_example\}}\ttnl
NOW SOLVE\ttnl
\textbf{\{l1\_name\}} word: \textbf{\{l1\_word\}}\ttnl
\textbf{\{l1\_name\}} context: \textbf{\{l1\_context\}}\ttnl
Clue: \textbf{\{clue\}}\ttnl
English word:\\
\end{raggedright}

\subsection{Difficulty}

\subsubsection*{Placeholders:}
\texttt{\{examples\}}: a block of examples for difficulty rating using examples from the training data. Examples with ratings converted to values close to 1, 3, and 5 are selected for each L1 separately.

\subsubsection*{Prompt template:}
\begin{raggedright}%
\noindent%
\ttfamily\footnotesize\microtypecontext{protrusion=false}%
You are an English language teacher teaching learners whose native language is \textbf{\{l1\_name\}}. Your task is to rate the difficulty of a vocabulary test item for native \textbf{\{l1\_name\}} speakers learning English.\ttnl

The test item consists of:\ttnl
- a \textbf{\{l1\_name\}} word,\ttnl
- a \textbf{\{l1\_name\}} context,\ttnl
- a clue indicating the first letter and word length of the English word,\ttnl
- the target English word, which is the only correct answer.\ttnl

Letter case does not matter. The learners are likely to respond with synonyms or misspellings to some items, but such responses are considered incorrect. Treat this as increasing the difficulty.\ttnl

Consider learners from beginner to advanced levels, weighting the intermediate learner most heavily. Rate how difficult the item is on a scale from 1 to 5:\ttnl
1 = very easy (almost everybody answers correctly)\ttnl
5 = very difficult (almost nobody answers correctly)\ttnl

Output exactly one digit (1, 2, 3, 4, or 5). Do not include any other text.\ttnl

\textbf{\{examples\}}\ttnl
\textbf{\{l1\_name\}} word: \textbf{\{l1\_word\}}\ttnl
\textbf{\{l1\_name\}} context: \textbf{\{l1\_context\}}\ttnl
Clue: \textbf{\{clue\}}\ttnl
English word: \textbf{\{en\_word\}}\ttnl
Difficulty:\\
\end{raggedright}

\onecolumn 
\section{Base Models and API Models}
\label{app:models}

\autoref{tab:base-models} lists base LLMs and MLMs that we fine-tuned in our experiments. \autoref{tab:prompt-models} lists LLMs that we used for prompt-based features via API.

\begin{table*}[!h]
\newcommand{\hfmodel}[1]{%
  \href{https://huggingface.co/#1}{\texttt{#1}}%
}%
\centering
\small\setlength{\tabcolsep}{5pt}
\begin{tabularx}{\linewidth}{llX}
\toprule
Model Name & Hugging Face Model ID & Systems/References to Results \\
\midrule
GLM-4-32B & \hfmodel{zai-org/GLM-4-32B-Base-0414} & \texttt{finetuned\_llms}$^\ast$ \\
Qwen2.5-32B & \hfmodel{Qwen/Qwen2.5-32B} & \texttt{finetuned\_llms}$^\ast$ \\
Ministral-3-14B & \hfmodel{mistralai/Ministral-3-14B-Base-2512} & \texttt{finetuned\_llms}$^\ast$; $\le$14B LLM Average;\newline
ablations in \autoref{tab:rmse-llm-ablation}
\\
Qwen2.5-14B & \hfmodel{Qwen/Qwen2.5-14B} & $\le$14B LLM Average \\
GLM-4-9B & \hfmodel{zai-org/glm-4-9b} & $\le$14B LLM Average; $\le$9B LLM Average \\
Qwen2.5-7B & \hfmodel{Qwen/Qwen2.5-7B} & $\le$9B LLM Average \\
Ministral-3-8B & \hfmodel{mistralai/Ministral-3-8B-Base-2512} & $\le$9B LLM Average \\
Qwen2.5-1.5B & \hfmodel{Qwen/Qwen2.5-1.5B} & model size comparison in \autoref{app:model-size} \\
Qwen2.5-0.5B & \hfmodel{Qwen/Qwen2.5-0.5B} & model size comparison in \autoref{app:model-size} \\
mmBERT-b (mmBERT-base) & \hfmodel{jhu-clsp/mmBERT-base} & \texttt{closed\_max}; regression and ablations in \autoref{tab:rmse-closed-models} \\
XLMR-b (XLM-RoBERTa-base) & \hfmodel{xlm-roberta-base} & baselines; regression in \autoref{tab:rmse-closed-models} \\
XLMR-l (XLM-RoBERTa-large) & \hfmodel{xlm-roberta-large} & regression in \autoref{tab:rmse-closed-models} \\
\bottomrule
\multicolumn{3}{l}{$^\ast$\parbox[t]{\linewidth-\widthof{$^\ast$}-4\tabcolsep}{The three models used in \texttt{finetuned\_llms}, were also used in the \texttt{finetuned\_llms\_plus} and \texttt{open\_max} submissions, and the $\le$32B LLM Average ensemble.}}\\
\end{tabularx}
\caption{Open-weight base models used in fine-tuning experiments.}\label{tab:base-models}
\end{table*}

\begin{table*}[!h]
\centering
\small\setlength{\tabcolsep}{5pt}
\begin{tabularx}{\linewidth}{lllXl}
\toprule
Model Name & Provider & Model ID (Snapshot) & Explainable Features & Additional Features \\
 & & & (\texttt{explainable}; \texttt{open\_max}) & (\texttt{open\_max}) \\
\midrule
GPT-5.2 & OpenAI & \texttt{gpt-5.2-2025-12-11} & Lexical ambiguity; Spelling difficulty; L1 calque & Difficulty \\
GPT-4.1 & OpenAI & \texttt{gpt-4.1-2025-04-14} &  & Trickiness; Difficulty \\
GPT-4.1-mini & OpenAI & \texttt{gpt-4.1-mini-2025-04-14} &  & Trickiness \\
GPT-4.1-nano & OpenAI & \texttt{gpt-4.1-nano-2025-04-14} &  & Trickiness \\
DeepSeek-V3.2 & DeepSeek & \texttt{deepseek-chat} & Lexical ambiguity &  \\
\bottomrule
\end{tabularx}
\caption{Models used for features based on LLM prompting.}\label{tab:prompt-models}
\end{table*}

\section{Hyperparameters}\label{app:hyper}

\autoref{tab:hyper} shows the general fine-tuning hyperparameters. For XLM-RoBERTa, we followed \citet{skidmore-etal-2025-transformer}. For other models, we performed a limited search for the learning rate and epochs using cross-validation. For mmBERT, a higher learning rate would likely be optimal, but 3e-5 was the largest we evaluated. For LLMs, we found that the learning rate of 1e-4 performed better than 1.5e-4, but we could not rerun the training for Qwen2.5-32B in time for the final submission. \autoref{tab:qlora} lists the quantization and LoRA hyperparameters used for all LLM fine-tuning.

For the XGBoost regressor (\texttt{XGBRegressor} in XGBoost's Scikit-Learn API\footnote{\url{https://xgboost.readthedocs.io/en/release_3.1.0/python/python_api.html\#xgboost.XGBRegressor}}), we set \texttt{max\_depth=3}, \texttt{learning\_rate=0.1}, and \texttt{n\_estimators=200}, while using default values for other hyperparameters.


\begin{table}[h]
\centering
\small\setlength{\tabcolsep}{5.2pt}
\begin{tabular}{lrrrrrr}
\toprule
Model & Epochs & Batch size & Grad. accum. & Learning rate & Weight decay & Warmup ratio \\
\midrule
XLM-RoBERTa (both) & 5 & 32 & --- & 3e-5 (linear) & 0.1 & 0.1 \\
mmBERT-base & 16 & 16 & --- & 3e-5 (const.) & 0.1 & 0.1 \\
Qwen2.5-32B & 4 & 2 & 8 & 1.5e-4 (const.) & --- & --- \\
All other LLMs & 4 & 2 & 8 & 1e-4 (const.) & --- & --- \\
\bottomrule
\end{tabular}
\caption{General fine-tuning hyperparameters used for different models.}\label{tab:hyper}
\end{table}

\twocolumn

\begin{table}[h]
\centering
\small\setlength{\tabcolsep}{5.2pt}
\begin{tabular}[t]{lll}
\toprule
Group & Hyperparameter & Value \\
\midrule
Quantization
  & Bit-width & 4-bit \\
  & Data type & NFloat4 \\
  & Double quant. & Yes \\
\midrule
QLoRA-related
  & Compute type & BFloat16 \\
  & Optimizer & \texttt{paged\_adamw\_8bit} \\
\midrule
LoRA adapter
  & Rank $r$ & 8 \\
  & Scaling $\alpha$ & 16 \\
  & Dropout rate & 0.05 \\
  & Target modules & All linear \\
  & Bias adaptation & No \\
\bottomrule
\end{tabular}
\caption{Quantization and LoRA hyperparameters used for all LLM fine-tuning experiments.}
\label{tab:qlora}
\end{table}

\section{Results Reported as Pearson's Correlation Coefficients}
\label{app:pcc}

\autoref{tab:pcc-open-models-ensembles}, \autoref{tab:pcc-closed-models}, \autoref{tab:pcc-llm-ablation}, and \autoref{tab:pcc-mlm-ablation} show Pearson's correlation coefficient (PCC) evaluation corresponding to the RMSE evaluation shown in the main text in Tables~\ref{tab:rmse-open-models-ensembles} to \ref{tab:rmse-mlm-ablation}.

\begin{table}[t]
\setlength{\tabcolsep}{3.1pt}
\footnotesize
\centering
\begin{tabularx}{\linewidth}{Xcccc}
\toprule
System & Chinese & German & Spanish & Mean \\
\midrule
\narrowtt{open\_max} & 0.927 & \textbf{0.916} & \textbf{0.920} & \textbf{0.921} \\
\narrowtt{finetuned\_llms\_plus} & \textbf{0.928} & 0.915 & 0.919 & 0.921 \\
\narrowtt{finetuned\_llms} & 0.925 & 0.914 & 0.915 & 0.918 \\
\midrule
$\le$32B LLM Average & 0.926 & 0.914 & 0.916 & 0.918 \\
- GLM-4-32B & 0.918 & 0.907 & 0.906 & 0.910 \\
- Qwen2.5-32B & 0.917 & 0.902 & 0.907 & 0.909 \\
- Ministral-3-14B & 0.915 & 0.902 & 0.907 & 0.908 \\
$\le$14B LLM Average & 0.921 & 0.905 & 0.908 & 0.911 \\
$\le$9B LLM Average & 0.916 & 0.901 & 0.900 & 0.906 \\
\midrule
Open-Track Baseline & 0.804 & 0.786 & 0.783 & 0.791 \\
Statistical Optimum & 0.989 & 0.990 & 0.995 & 0.991 \\
\bottomrule
\end{tabularx}
\caption{PCC of our open-track submissions, compared with average ensembles by model size, individual models, and the shared task's open-track baseline.}
\label{tab:pcc-open-models-ensembles}
\end{table}

\begin{table}[t]
\setlength{\tabcolsep}{3.1pt}
\footnotesize
\centering
\begin{tabularx}{\linewidth}{Xcccc}
\toprule
System & Chinese & German & Spanish & Mean \\
\midrule
\texttt{closed\_max} & \textbf{0.874} & \textbf{0.844} & \textbf{0.854} & \textbf{0.857} \\
\texttt{explainable} & 0.837 & 0.779 & 0.789 & 0.802 \\
\texttt{traditional} & 0.767 & 0.747 & 0.721 & 0.745 \\
\midrule
\texttt{exp.}: std. inference & 0.820 & 0.768 & 0.776 & 0.788 \\
\texttt{exp.}: lin. regression & 0.815 & 0.766 & 0.769 & 0.783 \\
\midrule
Closed-Track Baseline & 0.753 & 0.773 & 0.765 & 0.764 \\
\bottomrule
\end{tabularx}
\caption{PCC of our closed-track submissions, compared with two variants of the \texttt{explainable} model and the shared task's closed-track baseline.}
\label{tab:pcc-closed-models}
\end{table}

\begin{table}[t]
\setlength{\tabcolsep}{3.1pt}
\footnotesize
\centering
\begin{tabularx}{\linewidth}{Xcccc}
\toprule
Method (Base Model) & Chinese & German & Spanish & Mean \\
\midrule
\makebox[0pt][l]{Ours (Ministral-3-14B)} & \textbf{0.915} & \textbf{0.902} & 0.907 & \textbf{0.908} \\
- single language & 0.909 & 0.890 & \textbf{0.907} & 0.902 \\
- out-of-language & 0.862 & 0.879 & 0.876 & 0.872 \\
- short prompt & 0.911 & 0.899 & 0.906 & 0.905 \\
- standard loss & 0.892 & 0.885 & 0.886 & 0.887 \\
- std.\ loss \& inference & 0.859 & 0.853 & 0.850 & 0.854 \\
\bottomrule
\end{tabularx}
\caption{PCC of ablations of our LLM-based model.}
\label{tab:pcc-llm-ablation}
\end{table}

\begin{table}[t]
\setlength{\tabcolsep}{3.1pt}
\footnotesize
\centering
\begin{tabularx}{\linewidth}{Xcccc}
\toprule
Method (Base Model) & Chinese & German & Spanish & Mean \\
\midrule
Ours (mmBERT-b) & 0.837 & \textbf{0.837} & \textbf{0.826} & \textbf{0.833} \\
- single language & \textbf{0.839} & 0.816 & 0.824 & 0.826 \\
- out-of-language & 1.115 & 1.102 & 1.193 & 1.136 \\
- short prompt & 0.834 & 0.827 & 0.825 & 0.828 \\
- standard loss & 0.807 & 0.794 & 0.788 & 0.796 \\
- std.\ loss \& inference & 0.779 & 0.772 & 0.776 & 0.776 \\
\midrule
Regression (XLMR-b) & 0.710 & 0.729 & 0.693 & 0.711 \\
Regression (XLMR-l) & 0.813 & 0.823 & 0.801 & 0.812 \\
Reg.\ (mmBERT-b) & 0.812 & 0.798 & 0.797 & 0.802 \\
\bottomrule
\end{tabularx}
\caption{PCC of ablations of our MLM-based model, compared with using a standard regression head and different base models (XLM-RoBERTa base/large).}
\label{tab:pcc-mlm-ablation}
\end{table}

\section{Comparison of Models Across Sizes}\label{app:model-size}

In \autoref{tab:rmse-all-llms} and \autoref{tab:pcc-all-llms}, we compare all base models we used in experiments, including models that are only reported in ensembles in the main text. To facilitate the comparison of models of different architectures (decoder LLMs and encoder MLMs), we also add Qwen models of sizes 1.5B and 0.5B, as well as XLM-RoBERTa models fine-tuned with the same prompt and method. We can observe that the model size, rather than the architecture, determines performance in this task. At the same size, however, MLMs are more compute-efficient.

All models were fine-tuned using the same cross-entropy loss with soft targets and the same prompt, utilizing hyperparameters listed in \autoref{app:hyper}.


\begin{table}[t]
\setlength{\tabcolsep}{3.1pt}
\footnotesize
\centering
\begin{tabularx}{\linewidth}{Xcccc}
\toprule
Base Model & Chinese & German & Spanish & Mean \\
\midrule
GLM-4-32B & \textbf{0.678} & \textbf{0.769} & 0.805 & \textbf{0.751} \\
Qwen2.5-32B & 0.678 & 0.777 & 0.799 & 0.752 \\
\midrule
Ministral-3-14B & 0.681 & 0.781 & \textbf{0.799} & 0.753 \\
Qwen2.5-14B & 0.701 & 0.801 & 0.832 & 0.778 \\
\midrule
GLM-4-9B & 0.744 & 0.858 & 0.930 & 0.844 \\
Ministral-3-8B & 0.723 & 0.809 & 0.835 & 0.789 \\
Qwen2.5-7B & 0.722 & 0.842 & 0.882 & 0.816 \\
\midrule
Qwen2.5-1.5B & 0.800 & 0.955 & 0.998 & 0.918 \\
\midrule
Qwen2.5-0.5B & 0.851 & 1.021 & 1.078 & 0.983 \\
XLMR-l (550B) & 0.924 & 1.001 & 1.095 & 1.007 \\
\midrule
mmBERT-b (307M) & 0.921 & 0.984 & 1.063 & 0.989 \\
XLMR-b (270B) & 1.042 & 1.136 & 1.220 & 1.133 \\
\bottomrule
\end{tabularx}
\caption{RMSE of all individual base LLMs and MLMs.}
\label{tab:rmse-all-llms}
\end{table}

\begin{table}[t]
\setlength{\tabcolsep}{3.1pt}
\footnotesize
\centering
\begin{tabularx}{\linewidth}{Xcccc}
\toprule
Base Model & Chinese & German & Spanish & Mean \\
\midrule
GLM-4-32B & \textbf{0.918} & \textbf{0.907} & 0.906 & \textbf{0.910} \\
Qwen2.5-32B & 0.917 & 0.902 & \textbf{0.907} & 0.909 \\
\midrule
Ministral-3-14B & 0.915 & 0.902 & 0.907 & 0.908 \\
Qwen2.5-14B & 0.909 & 0.895 & 0.897 & 0.900 \\
\midrule
GLM-4-9B & 0.900 & 0.880 & 0.874 & 0.885 \\
Ministral-3-8B & 0.904 & 0.895 & 0.898 & 0.899 \\
Qwen2.5-7B & 0.904 & 0.884 & 0.884 & 0.890 \\
\midrule
Qwen2.5-1.5B & 0.883 & 0.848 & 0.848 & 0.860 \\
\midrule
Qwen2.5-0.5B & 0.863 & 0.824 & 0.820 & 0.836 \\
XLMR-l (550B) & 0.835 & 0.832 & 0.816 & 0.827 \\
\midrule
mmBERT-b (307M) & 0.837 & 0.837 & 0.826 & 0.833 \\
XLMR-b (270B) & 0.786 & 0.775 & 0.761 & 0.774 \\
\bottomrule
\end{tabularx}
\caption{PCC of all individual base LLMs and MLMs.}
\label{tab:pcc-all-llms}
\end{table}

\section{Simulation of a Statistical Optimum}
\label{app:stat-opt}

The difficulty scores provided by the KVL data are based on test item responses. The number of responses collected for each item (120 to 228 responses) determines the precision of the data, which can be expressed as a confidence interval. \citet[Sec.~6.3, App~5]{schmitt_etal_2024_knowledge} provide 83\% confidence intervals for selected bands of the complete KVL data (e.g., items ranked 200--299 for L1=Spanish) expressed in numbers of ranks (e.g., 44). The rationale for 83\% is that it corresponds to a significance level of 5\% for differences in pairwise ordering (e.g., there is no significant difference in difficulty between items within ranks 200 and 44). The widest confidence intervals are 69, 95, and 108 ranks for Spanish, Chinese, and German, respectively. While the span of 100 ranks is given as a rule-of-thumb criterion for ``strong confidence'', we use the above per-L1 maximum widths $w$ for our simulation.

We simulate our statistical optimum predictions by taking the most distant difficulty in the complete KVL Data within $\pm w$ ranks of the predicted data point and using it as a prediction. In accordance with the originally reported confidence intervals, this process is based on the complete KVL data (training, development, and test subsets), although we predict only for the test data. Such predictions could be considered to have no statistically significant difference from the gold standard data; hence, we report their RMSE and correlation as a ``statistical optimum''.

\onecolumn

\section{Predicting in the Probability Space}\label{app:probspace}

As shown in \autoref{tab:probspace}, fine-tuning an LLM to predict in the probability space instead of the logit space, in which the GLMM scores are, resulted in a decrease in performance. We therefore used the raw GLMM scores for the main experiments.

\begin{table*}[h]

\setlength{\tabcolsep}{3.1pt}

\footnotesize

\centering

\begin{tabular}{lcccccccc}

\toprule

& \multicolumn{4}{c}{$\downarrow$ RMSE} & \multicolumn{4}{c}{$\uparrow$ PCC} \\

\cmidrule(lr){2-5} \cmidrule(lr){6-9}

Target Values & Chinese & German & Spanish & Mean & Chinese & German & Spanish & Mean \\

\midrule

Logits: GLMM score

& \textbf{0.681} & \textbf{0.781} & \textbf{0.799} & \textbf{0.753}

& \textbf{0.915} & \textbf{0.902} & \textbf{0.907} & \textbf{0.908} \\

Probabilities: expit(GLMM score)

& 0.738 & 0.853 & 0.877 & 0.822

& 0.905 & 0.889 & 0.895 & 0.896 \\

\bottomrule

\end{tabular}

\caption{Performance of Ministral-3-14B when fine-tuned on the raw GLMM score (logits), as we did in all experiments, compared with the same model fine-tuned on the corresponding probabilities. In both cases, the values were linearly transformed to the 1-to-5 difficulty scale we use in our prompt, and all other parameters were the same. RMSE and PCC were measured for GLMM score values in the logit space for both.}

\label{tab:probspace}

\end{table*}

\end{document}